\newif\ifpreprint
\newif\ifjournalsubmission
\newif\iffinal

\preprinttrue

\ifpreprint
    \documentclass{iopjournal}
\else
    \ifjournalsubmission
        \documentclass[anonymous]{iopjournal}
    \else
        \documentclass{iopjournal}
    \fi
\fi
\usepackage{microtype}
\usepackage[nolist]{acronym}
\begin{acronym}

\acro{OSL}[OSL]{Optically Stimulated Luminescence}
\acro{TLD}[TLD]{Thermoluminescence Dosimeter}
\acro{APD}[APD]{Active Personal Dosimeter}
\acro{EPD}[EPD]{Electronic Personal Dosimeter}
\acro{MCS}[MCS]{Monte Carlo Simulation}
\acro{DSR}[DSR]{Dose Structured Report}
\acro{CT}[CT]{Computed Tomography}
\acro{MRI}[MRI]{Magnetic Resonance Imaging}
\acro{CG}[CG]{Computer Graphics}
\acro{CNN}[CNN]{Convolutional Neural Network}
\acro{PRT}[PRT]{Precomputed Radiance Transfer}
\acro{NeRF}[NeRF]{Neural Radiance Field}
\acro{MLP}[MLP]{Multilayer Perceptron}
\acro{CUDA}[CUDA]{Compute Unified Device Architecture}
\acro{PACS}[PACS]{Picture Archiving and Communication System}
\acro{IR}[IR]{Interventional Radiology}
\acro{VR}[VR]{Virtual Reality}
\acro{AR}[AR]{Augmented Reality}
\acro{API}[API]{Application Programming Interface}
\acro{FCNN}[FCNN]{Fully Connected Neural Network}
\acro{CNN}{Convolutional Neural Network}
\acro{RAF}{Realistic Anthropomorphic Flexible}
\acro{NVS}{Novel View Synthesis}
\acro{LSTM}{Long-Short-Term-Memory}
\acro{RNN}{Recurrent Neural Network}
\acro{SMAPE}{Symmetric Mean Absolute Percentage Error}
\acro{GPR}{Gamma Passing Ratio}
\acro{IoU}{Intersection over Union}
\acro{SSIM}{Structural Similarity Index Measure}
\acro{MADD}{Maximum Allowed Dose Difference}
\acro{FiLM}{Feature-wise Linear Modulation}
\acro{GMU}{Gated Multimodal Unit}
\acro{SiLU}{Sigmoid Linear Unit}

\acro{ICG}[ICG]{Institute of Computer Graphics of the Technical University Braunschweig}
\acro{PTB}[PTB]{Physikalisch-Technische Bundesanstalt}
\acro{ECAP}[ECAP]{Erlangen Centre for Astroparticle Physics}
\acro{SCK}[SCK CEN]{Belgian Nuclear Research Centre}

\acro{WP}[WP]{Work Package}

\acro{PBRFNet}{Parametric Beam Radiation Field Network}
\acro{SPERFNet}{Spectral Enhanced Radiation Field Network}
\acro{SRBFNet}{Static Rotatable Beam Field Network}

\end{acronym}

\usepackage{hyperref}
\usepackage[english]{babel}
\usepackage{csquotes}
\usepackage{graphicx}
\usepackage{wrapfig}
\usepackage{esvect}
\usepackage{tikz}
\usetikzlibrary{
    positioning,
    arrows.meta,
    fit,
    calc,
    shapes.geometric
}
\usepackage{amsmath}
\usepackage{svg}
\usepackage{gensymb}
\usepackage[a4paper]{geometry}
\usepackage{booktabs}
\usepackage{amssymb}
\usepackage{multirow}
\usepackage{calrsfs}
\DeclareMathAlphabet{\pazocal}{OMS}{zplm}{m}{n}

\usepackage{siunitx}
\usepackage[capitalize]{cleveref}
\usepackage{fontawesome5}
\usepackage{hyperref}

\crefname{paragraph}{Sec.}{Secs.}
\crefname{section}{Sec.}{Secs.}
\Crefname{section}{Section}{Sections}
\Crefname{table}{Table}{Tables}
\crefname{table}{Tab.}{Tabs.}
\Crefname{figure}{Figure}{Figures}
\crefname{figure}{Fig.}{Figs.}
\Crefname{chapter}{Chapter}{Chapters}
\crefname{chapter}{Ch.}{Chs.}

\providecommand{\eg}[0]{\emph{e.g.}, }

\providecommand{\etal}[0]{\emph{et~al.\ }}

\usepackage[colorinlistoftodos]
{todonotes}
\definecolor{darkorange}{rgb}{1.0, 0.55, 0.0} 
\definecolor{LightCyan}{rgb}{0.88,1,1}
\definecolor{darkred}{rgb}{0.7, 0.0, 0.0}
\definecolor{forrestgreen}{rgb}{0, 0.36, 0.0}

\tikzset{
    block/.style={
        draw,
        rectangle,
        rounded corners,
        minimum height=0.5cm,
        minimum width=1.6cm,
        align=center,
        font=\small
    },
    smallblock/.style={
        draw,
        rectangle,
        rounded corners,
        minimum height=0.4cm,
        minimum width=1.2cm,
        align=center,
        font=\scriptsize
    },
    input/.style={
        draw,
        trapezium,
        trapezium left angle=70,
        trapezium right angle=110,
        minimum height=0.5cm,
        minimum width=1.6cm,
        align=center,
        font=\small
    },
    fuse/.style={
        draw,
        diamond,
        aspect=2,
        inner sep=1pt,
        align=center,
        font=\scriptsize
    },
    arrow/.style={
        ->,
        thick
    }
}

\newcommand{\fluence}{\ensuremath{\Phi^{3}_{\mathrm{\gamma}}}~}

\def\FINAL{}
\usepackage[normalem]{ulem}
\definecolor{yellow-green}{rgb}{0.4, 0.53, 0.14}
\newcommand{\rev}[1]{\ifx&#1&\else\colorbox{teal!20}{#1}\fi}
\newcommand{\added}[2][]{\ifdefined\FINAL#2\else\rev{#1} \textcolor{yellow-green}{\textbf{#2}}\fi}
\newcommand{\deleted}[2][]{\ifdefined\FINAL\else\rev{#1} \textcolor{red}{\sout{#2}}\fi}
\newcommand{\replaced}[3][]{\ifdefined\FINAL#3\else\rev{#1} \deleted{#2} \added{#3}\fi}

\begin{document}

\articletype{Paper} 

\title{Learning-Based Estimation of Spatially Resolved Scatter Radiation Fields in Interventional Radiology}

\author{Felix Lehner$^{1,3}$\orcid{0000-0001-9748-183X}, Pasquale Lombardo$^{2}$\orcid{0000-0001-7095-7295}, Susana Castillo$^{3,5}$\orcid{0000-0003-1245-4758}, Oliver Hupe$^{1}$\orcid{0000-0001-6561-3375}, and Marcus Magnor$^{3,4,5}$\orcid{0000-0003-0579-480X}}

\affil{$^1$Physikalisch-Technische Bundesanstalt (PTB), Braunschweig, Germany}

\affil{$^2$Belgian Nuclear Research Centre (SCK CEN), Boeretang, Mol, Belgium}

\affil{$^3$Institute for Computer Graphics, Technical University Braunschweig, Braunschweig, Germany}

\affil{$^4$Physics and Astronomy, University of New Mexico, New Mexico, USA}

\affil{$^5$Cluster of Excellence PhoenixD, Leibniz University Hannover, Hannover, Germany}

\email{lehner@cg.cs.tu-bs.de, felix.lehner@ptb.de}

\keywords{\acl{IR}, \acl{MCS}, \acs{FCNN}, U-Net, Dataset, X-ray}

\ifpreprint
\vspace{0.5cm}
\href{https://github.com/Centrasis/radfield3d-nn}{\textcolor{black}{\faGithub}}
\thickspace
\thickspace
\thickspace
\href{https://box.ptb.de/getlink/fi5etj6QwfD5PFgNqbaSo1VP/JML-2025}{\textcolor{black}{\faDatabase}}
\fi

\begin{abstract}
\replaced[R1C1, R1C12, R1C14]{We present an in-depth analysis on how to build and train neural networks to estimate the spatial distribution of scattered radiation fields for radiation protection dosimetry in medical radiation fields, such as those found in interventional radiology and cardiology.
We present three different synthetically generated datasets with increasing complexity for training, using a \acl{MCS} application based on Geant4. On those datasets, we evaluate convolutional and fully connected architectures of neural networks to demonstrate which design decisions work well for reconstructing the fluence and spectra distributions over the spatial domain of such radiation fields. All our datasets, as well as our training pipeline, are published as open source in separate repositories.
}{We present three variants of a lightweight, fully connected artificial neural network, suited for interactive estimation of three-dimensional, spatially resolved volumes of scattered radiation fields and a corresponding training pipeline for radiation protection dosimetry in medical radiation fields, such as those found in interventional radiology and cardiology. Accompanying, we present three different synthetically generated datasets with increasing complexity for training, generated using RadField3D, a Monte Carlo simulation application based on Geant4. As the primary scatter object, we employed the torso of a male Alderson RANDO phantom. On those datasets, we evaluate convolutional and fully connected architectures of neural networks to demonstrate which design decisions work well for reconstructing the fluence and spectra distributions over the spatial domain of such radiation fields. All our datasets, as well as our training pipeline, are published as open source in separate repositories.\\
To evaluate the presented neural networks, we define and assess several metrics. Across these measures, the model variants demonstrate good spatial agreement between predicted and ground-truth radiation fields, particularly within specific
regions of interest within the radiation field. Of particular relevance for potential application in out-of-field dosimetry is the SMAPE of the scatter radiation field, which represents the most challenging metric and was consistently above \(84\,\%\).
}
\end{abstract}

\section{Introduction}
The spatial distribution of radiation in a volume is of interest for various applications in radiation protection and occupational dosimetry. Knowledge gained from the analysis of irradiation scenarios can be used for optimizing radiation exposure. This is important for work, for example, in nuclear power plants, radioactive waste disposal, or in radiological medicine, such as~\ac{IR}, in which medical staff performs surgeries under fluoroscopy. The current state of the art relies on the utilization of personal dosemeters, which are sufficient for spatially homogeneous radiation distributions, but inherently misestimate personal doses in inhomogeneous radiation fields.

In the context of \ac{IR}, medical staff are exposed to inhomogeneous radiation fields due to their proximity to the patient, which complicates the accurate assessment of individual doses. The reliability of current personal dosimetry methods is contingent upon the uniform distribution of radiation fields. However, this assumption of uniformity does not hold in \ac{IR} settings. To address this issue, previous research has proposed the use of computational dosimetry systems~\cite{oconnor_feasibility_2022}, which are designed to monitor the locations and postures of all individuals involved in an \ac{IR} procedure. In order to assess doses in these exposure scenarios, such systems make use of currently available methods, that are mainly based on \acp{MCS}~\cite{chatzisavvas_monte_2021}.

Nonetheless, the existing \acp{MCS} of radiation transport lack the necessary speed for real-time dose calculations, a limitation that also affects the software used for \ac{VR} or \ac{AR} training. \ac{VR} training systems can play a pivotal role in diminishing radiation exposure during interventions by enhancing the radiation awareness of medical staff, as evidenced by at least two research projects~\cite{rainford_student_2023, fujiwara_virtual_2024}. Notwithstanding their significance, there is a paucity of  reliable and realistic radiation dosimetry data to be used in real-time visualization with these systems.

To advance radiation protection, it is necessary to develop accelerated methods for the simulation of radiation transport. In this research work, we explore the surrogate model approach for \ac{MCS} acceleration, whereby we learn spatially resolved radiation field distributions from a \acl{MCS}. Subsequently, the neural networks serve as efficient models for radiation-field reconstruction, providing a real-time capable radiation-transport simulation method. Furthermore, we have generated and made available a variety of datasets for the training of our models and analogous ones. The generation of our datasets was facilitated by the extension and utilization of the RadField3D~\cite{lehner_radfield3d_2025} simulation software which itself is built upon the Geant4~\cite{agostinelli_geant4simulation_2003} \ac{MCS} framework to ensure physical correctness, as Geant4 is well established and validated for the targeted use-case.

Our primary innovation is the creation of three datasets of spatially resolved, three-dimensional fluence and spectra distributions for the training and evaluation of neural networks for the reconstruction of radiations fields for~\ac{IR} procedures. We accompany these with concrete neural network implementations and design recommendations based on ablation studies. Additionally, we define several metrics useful for evaluating and comparing spatially resolved radiation field estimators.
\subsection{Monte-Carlo Simulation in Radiation Protection}~\label{sec:MCS-RadProt}
The technique of using \ac{MCS} to calculate the radiation transport of complex measurements is already widely spread in various disciplines of dosimetry, from radiation protection to medical physics. Therefore, a set of well-established and tested general-purpose \ac{MCS} toolkits for radiation transport already exists, such as Geant4~\cite{agostinelli_geant4simulation_2003}, MCNP~\cite{werner_mcnp_2018}, and EGSnrc~\cite{canada_egsnrc_2021}. As these toolkits are meant for general purpose applications ranging from astrophysics over nuclear reactor physics to dosimetry calculations, fine-tuning the \ac{MCS} parameters for each specific use case is required. There are frameworks built upon these toolkits, such as GATE~\cite{jan_gate_2004}, which is based on Geant4, and it has already been used for assessing doses to medical staff in nuclear medicine by tailoring simulation models based on the real tracking of personnel movements \added[R1C13]{as demonstrated by Rondón et al.}~\cite{rondon:2025}.
However, as highlighted by Rondón et al., current \ac{MCS} capabilities are still far from real-time as they require, at best, several tens of seconds if not minutes to be performed. This is particularly challenging in \ac{IR}, as the calculation of radiation-fluence distributions in the space around the patient during interventions is computationally demanding and can require hours of simulations. However, real-time simulation capabilities are becoming increasingly necessary for computational radiation protection dosimetry of staff, as proposed by prior research projects, such as the PODIUM project~\cite{oconnor_feasibility_2022}.

Besides the real-time dosimetry, the \ac{AR} and \ac{VR} applications for training interventional radiologists and associated personnel demand even faster execution speeds. Nevertheless, especially in \ac{AR} and \ac{VR}, we can trade absolute accuracy for speed within reasonable bounds. These applications have already been presented at various times in the last years, for example in 2012 by a c-arm training extension for VirtX~\cite{virtx} or by the perceptual study of Rainford~\etal in 2023~\cite{rainford_student_2023}. All current implemented systems would benefit from the display of physically correct radiation fields during training, but at the moment these systems are only able to render radiation fields generated by simplified deterministic algorithms due to strict timing requirements. 
For these applications, achieving frame rates of at least \(90\) to \(120\,\mathrm{fps}\) requires total rendering times per volume of approximately \(8\) to \(11\,\mathrm{ms}\) to avoid perceptible visual latency~\cite{wang_effect_framerate_VR_2023}.

\subsection{Computational Personal Dosimetry}
Real-world measurements in \ac{IR} scenarios employing \acp{APD} have already revealed the complexity of contemporary radiation fields and high local dose rates that professionals are exposed to~\cite{hupe_determination_2011}. These measurements also implied that the definition of a representational location for wearing a personal dosemeter on the body of the medical staff is not possible, and therefore, it is likely to both under- or overestimate the received doses by the staff. In the course of the PODIUM project, the researchers developed a camera-based tracking system that was integrated with a \ac{MCS} application based on the MCNP~\cite{werner_mcnp_2018} toolkit. The simulation divided the room into frustums projected from the surface of a sphere placed around the patient at the isocenter of the scene~\cite{abdelrahman_first_2020, vanhavere_d9121_2020}. The evaluation of each event was conducted as it traversed one of the surface segments of the sphere, effectively projecting it along the normal of that surface segment through the scene. The validation of the simulation was based on these frustums, with measurements of \acp{APD} worn by medical personnel during real interventions serving as the ground-truth. These interventions involved tracking the locations and postures of staff members.

\subsection{Hybrid Methods and Surrogate Models for Radiation Transport \ac{MCS}}
Due to its notable slowness, as evidenced in the research by Vanhavere \& Van Hoey~\cite{vanhavere_advances_2022}, the \ac{MCS} is ill-suited for real-time or even interactive applications. To address this issue, recent studies have explored the prospect of enhancing its calculation speed. These efforts entail the replacement of the most computationally inefficient components of the radiation transport simulations, which is referred to as a hybrid approach. These hybrid methods can be composed of denoising algorithms. The most promising results in this field are delivered by deep learning methods, that use artificial neural networks for transforming the noisy results of an early aborted \ac{MCS} to those of a fully executed \ac{MCS}, but with shorter execution times~\cite{raayai_ardakani_framework_2022, bai_deep_2021, peng_mcdnet_2019}. Although this approach has been the subject of more extensive research, it exhibits a significant disadvantage, as the \ac{MCS} needs a certain minimum of statistics to be denoise-able which inevitably comes at the price of a lower processing speed compared to a single neural network. Consequently, it does not meet the requirements for real-time applications especially in the context of \ac{VR} and \ac{AR} applications previously mentioned.

Other approaches focus on the complete replacement of the \ac{MCS} algorithm, with artificial neural networks. This innovative approach is called surrogate models and its feasibility was already demonstrated in a few studies for different scenarios~\cite{arias_radiation_2023, ye_research_2024, yisheng_validation_2023, Wang2025ResearchO3}. Most of these studies were conducted in the context of nuclear power plants, thus those methods aim to predict neutron radiation fields induced by nuclear isotopes. Therefore, those works were heavily focused on modeling shielding and its positioning regarding a static radiation source. For this use case, Wang~\etal proposed the use of a residual \ac{FCNN} for predicting neutron radiation fields under various shielding configurations~\cite{Wang2025ResearchO3}. As they could assume a static radiation source, they split the room into regions with similar conditions regarding the radiation field and trained multiple networks for each region. Concurrently, Ye~\etal implemented a 3D-\ac{CNN} for 3D reconstruction in a similar scenario~\cite{ye_research_2024}. This 3D-\ac{CNN} takes the unshielded radiation fields and shielding voxel geometry, convolutes those two channels down to a flat vector which is then rearranged as a 3D radiation field after the shielding process. Another use case for \ac{FCNN} is demonstrated in the research work of Yisheng~\etal\cite{yisheng_validation_2023}, where a \ac{FCNN} is used to predict the spatially resolved photon flux in a room based on the data of a single detector at a known position. The output of the neural network is a low resolution grid with $2000$ voxels. An example of the use of neural networks in the domain of this research work in the context of~\ac{IR} was provided by the research work of M. Villa Arias~\cite{arias_radiation_2023}. Here, a 3D-UNet is used to transform a CT-Scout-Scan to the surrounding radiation field around the patient induced by the CT-scanner.

\added[R1C2, R1C9, R2C8]{This work aims to contribute to the corpus of existing surrogate model approaches by focusing on predicting dynamic radiation fields using lightweight \acp{FCNN}. This enables the prediction of partial radiation fields within the region of interest, thereby supporting real-time capable radiation field estimation and, in theory, also allowing for higher resolutions of radiation fields. In contrast to prior approaches, our approach additionally provides the local energy spectra, making it possible to use those networks for correcting the energy dependence of real world dosemeters. In the following, we present \acp{FCNN} inspired by the architecture of the networks used for \acp{NeRF}~\cite{mildenhall_nerf_2020}.
As these networks model geometry implicitly, and our datasets do not incorporate variations in patient geometry, a direct comparison with the work of Arias~\etal~\cite{arias_radiation_2023} is neither meaningful nor feasible, given that their approach explicitly encodes patient geometry.}

\section{Materials and Methods}
As there is a lack of publicly available, and in this context, usable datasets our first contribution is the generation and provision of three datasets for training spatially resolved dose rate predicting neural networks.
For these three datasets, we implemented two distinct neural network architectures with additional variants for each, in order to analyze the influence of these architectures and specific hyperparameters on the quality of the predictions.
For this purpose, we considered two basic architectures: an \ac{FCNN} architecture inspired by \acp{NeRF}~\cite{mildenhall_nerf_2020} and a 3D-U-Net~\cite{cicek_3d_2016} architecture which is based on the \ac{CNN} architecture, but with an extra transposed path for reconstructing all voxels at once.

\subsection{Datasets}
We generated and provided three synthetic datasets of radiation fields for different situations with increasing parameter spaces. The datasets were all generated using the RadField3D~\cite{lehner_radfield3d_2025} dataset generator. \added[R1C3]{Per field, the generation took about \(20\,\mathrm{min}\) on our AMD Epyc 9474F 48-core CPU.} The resulting datasets consist of one radiation field per fixed set of parameters. These fields are stored in the RadFiled3D file format and contain three channels. One for the photons of the direct X-ray beam, one for the resulting scattered fields and one containing the binary voxel occupation of the modeled geometry. Each of the first two channels contains the following information per voxel: The simulated volumetric \textit{photon fluence} \fluence (fluence per cubic volume of a voxel), the volumetric \textit{photon energy distribution} \(p(E_\mathrm{\gamma})\) and the \textit{statistical error} \(\epsilon_{rel}\) as described in the research work about the \ac{MCS} application RadField3D. Additionally, all input parameters, such as the X-ray tube output spectra or the tubes position and direction, are stored per field.

\paragraph{\textbf{DS-01}:} The first dataset we generated, contains \(1250\) radiation fields. Each field has a resolution of 50\texttimes50\texttimes50 voxels with each voxel having an extent of \(2\,\mathrm{cm}\) in every dimension resulting in a total field extent of \(1\,\mathrm{m}\) per dimension. The varying parameter in this first dataset is the direction of the beam with \(\phi, \theta \in [0, 2\pi)\), while the X-ray tube output is fixed to a reference radiation quality of H-100 according to ISO 4037-1~\cite{iso_iso_2019} and the beam shape is fixed to a cone beam with an opening angle of \(\angle_{\mathrm{Beam}} = 10\degree\). The distance from the X-ray tube relative to the origin is \(2.5\,\mathrm{m}\). At the origin of the field, the headless torso of a male Alderson RANDO phantom\added{~\cite{alderson_rando_phantom}} is placed to act as the scatter object for the radiation beam. This setting is derived from the validation measurements conducted for the RadField3D data generator~\cite{lehner_radfield3d_2025}, and, although simplified, it is considered realistic for modeling this first exposure scenario.

\paragraph{\textbf{DS-02}:} The second dataset contains \(2156\) radiation fields and extends the previous one by the radiation source spectra as a varying parameter. Here we used random, but realistic, spectra for modeling primary radiation in the context of \ac{IR}. The range of the parameters was defined based on empirical feedback from medical physicist and radiologists, mainly gained during the PODIUM project.  Therefore, we defined an energy range for the tube output \added[R1C5]{peak} energy \(E_{\mathrm{tube}} \in [40, 125]\,\mathrm{keV}\). Further, we added aluminum with a random thickness \(t_{\mathrm{Al}} \in [2.5, 7.5]\,\mathrm{mm}\) and, optionally, copper, with a random-thickness \(t_{\mathrm{Cu}} \in [0.0, 0.9]\,\mathrm{mm}\) for simulating realistic filtering material used to improve the X-ray beam quality. As the anode material, we set tungsten, which is a standard reference material in \ac{IR} imaging devices, and used a random anode angle of \(\angle_{\mathrm{Anode}} \in [8, 12]\degree\). From that parameter space for the radiation source, we sampled sets of parameters and passed them to SpekPy~\cite{poludniowski_technical_2021} to generate reasonable X-ray spectra for the X-ray beams in this dataset.

\paragraph{\textbf{DS-03}:} In the third and most complex dataset, dynamic X-ray tube distancing is introduced. A total of \(3779\) distinct radiation fields were generated.  For this dataset, we \added{mostly} carried over the same base parameter space as described in the preceding datasets, \replaced{and extended it by sampling}{but in contrast to the previous datasets, we sampled} the X-ray tube distance from the scenes isocenter as \(d_{\mathrm{tube}} \in [35, 75]\,\mathrm{cm}\). In order to make the dataset even more realistic in representing typical real-world scenarios, the X-ray beam was collimated to a rectangular shape, with fixed dimensions of \(40\,\mathrm{cm} \times 30\,\mathrm{cm}\) in the plane passing through the origin \added{and thus intersecting the phantom}.

\paragraph{}
\begin{table}[!h]
    \centering
    \caption{Listing of the statistical errors, with respect to the fluence, from the scattered radiation fields separated by dataset. \(\mathrm{mean}(error)\) refers to the mean over all voxels in the dataset, while \(\mathrm{median}(error)\) refers to the mean over the median statistical error of each file inside the dataset.}
    \begin{tabular}{ccc}
        Dataset & \(\mathrm{mean}(error)\) & \(\mathrm{median}(error)\) \\ \hline
        DS-01 & \(9.7\,\%\) & \(0.03\,\%\) \\
        DS-02 & \(9.9\,\%\)  & \(0.03\,\%\) \\
        DS-03 & \(9.9\,\%\)  & \(0.03\,\%\) \\
    \end{tabular}\label{tbl:radfiled3D_stat_errors}
\end{table}
\added[R1C2]{RadField3D uses the variation of the relative voxel fluences during the simulation process to measure the statistical error per voxel. Those statistical errors from the scattered radiation field layer are evaluated continuously to determine the current overall statistical error. Once \(90\,\%\) of those voxels exhibit a statistical error equal to or less than \(10\,\%\), the simulation is aborted. This behavior is reflected by~\cref{tbl:radfiled3D_stat_errors} where we show the mean and median statistical errors present in our previously described datasets, as calculated by RadField3D.}

\paragraph{}
\added[R1C6, R1C8]{An example of the spatial distribution of relative air kerma rates in air \(\dot{K}_{\mathrm{air}}\) can bee seen in~\cref{fig:3d-projection-ds03}. The two projections were generated from a random field taken from DS-03 by excluding all doserates below \(0.5\,\%\) of the maximum doserate for improved visual understanding.
\begin{figure}[!h]
    \centering
    \includegraphics[width=0.95\linewidth]{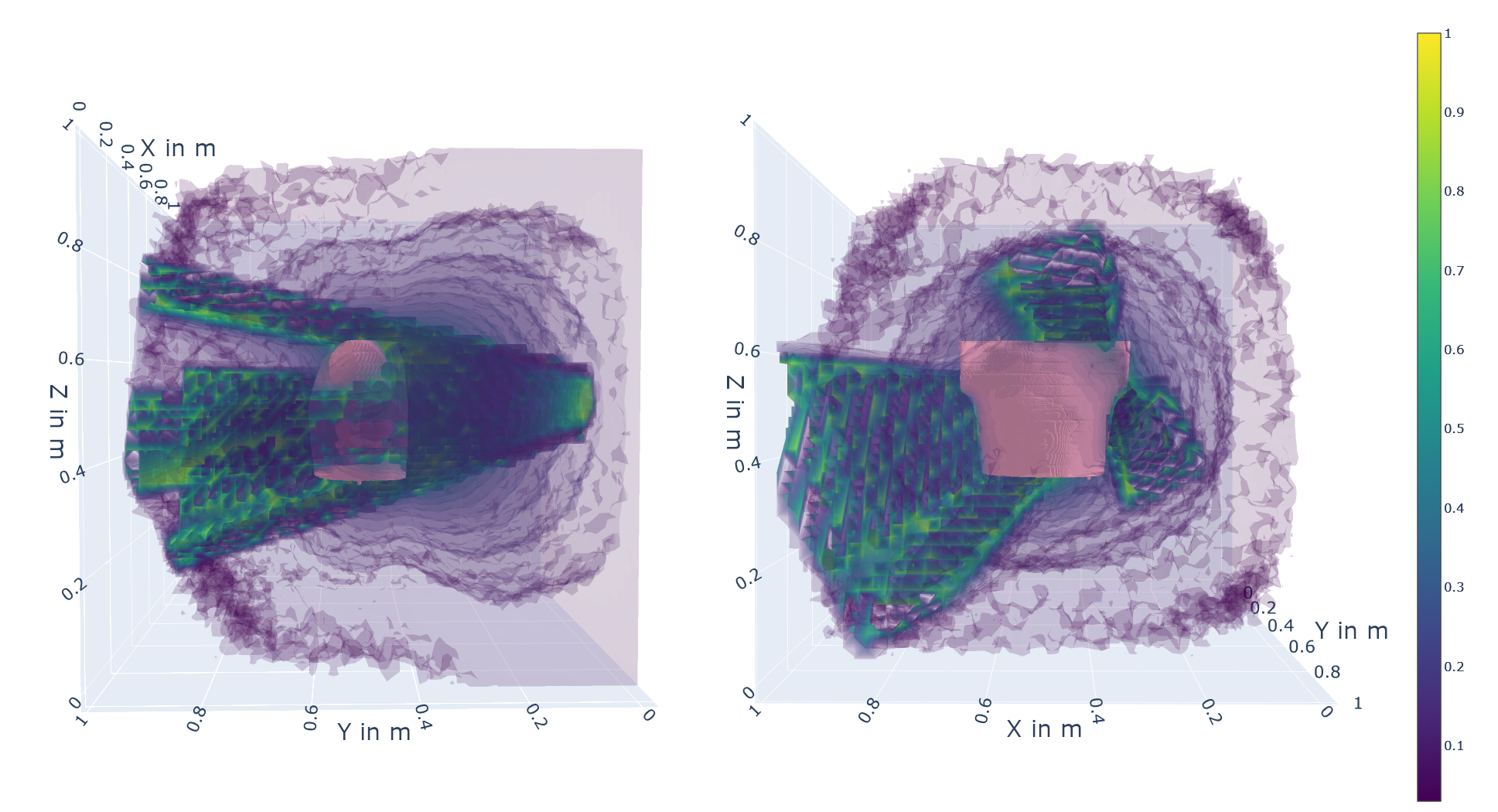}
    \caption{Two perspective projections of a random radiation field from DS-03: One projection along the X-axis (left) and one along Y-axis (right). The spatially resolved relative doserate \(\dot{K}_{\mathrm{air}}\) (air kerma in air) is color coded from transparent (\(<0.5\,\%\)) over blue (\(0.5\,\%\)) to yellow (\(100\,\%\)). The used scatter geometry (Male Alderson RANDO phantom torso) is colored in red.}
    \label{fig:3d-projection-ds03}
\end{figure}
}

\paragraph{}
In~\cref{tab:tbl_datasets}, we have summarized the distribution of each input parameter inside the three datasets and additionally the distribution of the contained fluences by providing the dynamic ranges \({DR}_{\mathrm{dB}}\) and the Gini coefficient \(G_{\mathrm{Gini}}\), which are defined as:
\begin{equation*}
{DR}_{\mathrm{dB}} = 10 \log_{10}\Bigl(\frac{\max(\fluence)}{\min(\fluence)}\Bigr)
\end{equation*}
\begin{equation*}
G_{\mathrm{Gini}} = \frac{1}{n^2\mu} \sum_{i=1}^nx_i(2i-n-1)
\end{equation*}
where \({DR}_{\mathrm{dB}}\) is used to describe the actual ranges needed for reconstruction and \(G_{\mathrm{Gini}}\) expresses the deviation of the present fluence distribution from a uniform distribution. For the calculation of \(G_{\mathrm{Gini}}\), the fluences are expected to be sorted in ascending order\replaced{.}{ , where \(\mu\) denotes the mean fluence and \(n\) the total count of voxels across the dataset. Finally,} a Gini coefficient of \(G_{\mathrm{Gini}} = 0\) indicates that the dataset is uniformly distributed, while \(G_{\mathrm{Gini}} = 1\) indicates that \added{all fluences of} the dataset \replaced{contains only one value or category}{are accumulated in one single bin of the fluence histogram}.

\begin{table}[h!]
    \caption{Statistics over the presented datasets including all varying input parameters.}~\label{tab:tbl_datasets}
    \centering
    \begin{tabular}{c|cccc|cc}
        Dataset & \(\overline{\angle}_{\mathrm{Beam}} \pm \sigma\) & \(\overline{dist}_{\mathrm{tube}} \pm \sigma\) & \(\overline{\mathrm{E}} \pm \sigma\) & \(\overline{\mathrm{keV_p}} \pm \sigma\) & \({DR}_{\mathrm{dB}}\) & \(G_{\mathrm{Gini}}\) \\
        \hline
        DS-01 & \(10.0\degree \pm 0.0\) & \(2.5\,\mathrm{m} \pm 0.0\) & \(80.7\,\mathrm{keV} \pm 0.0\) & \(100.0\,\mathrm{keV} \pm 0.0\) & \(43.9\,\mathrm{dB}\) & \(0.790\) \\
        DS-02 & \(10.0\degree \pm 0.0\) & \(2.5\,\mathrm{m} \pm 0.0\) & \(48.7\,\mathrm{keV} \pm 9.3\) & \(82.1\,\mathrm{keV} \pm 24.6\) & \(53.5\,\mathrm{dB}\) &\(0.872\) \\
        DS-03 & \(-\) & \(0.6\,\mathrm{m} \pm 0.1\) & \(48.7\,\mathrm{keV} \pm 9.3\) & \(82.1\,\mathrm{keV} \pm 24.6\) & \(76.0\,\mathrm{dB}\) & \(0.991\) \\
        \hline
    \end{tabular}
\end{table}

By observing the differences between our three datasets, one can see, that by reducing the volume of the direct beam inside the radiation field volume, the dynamic range \({DR}_{\mathrm{dB}}\) is drastically increased. In the case of DS-03 when smaller opening angles of the beam and smaller distances of the X-ray tube from the isocenter are used, particle fluences \fluence span over nearly \(8\) orders of magnitude, instead of the spanning of about \(5\) orders of magnitude for the other datasets. Concurrently, the distribution of fluences is more challenging, as the amount of voxels with fluences below \(0.1\,\%\) of the maximum fluence, rises by a factor of approximately \(253\) from \(0.3\,\%\) to \(76\,\%\).

\added[R1C3]{Overall, for all three datasets, the dataset generator RadField3D ensures, that \(90\,\%\) of all voxels do not exceed a statistical error of \(10\,\%\) with respect to fluence.}

\replaced[R1C7, R1C12, R2C7]{
Removed section about different data normalization mappings.
}{
\paragraph{\textbf{Data normalization:}} All datasets used are provided normalized by RadField3D. This applies both to the voxelwise photon energy spectra, which are probability distributions with unit integral, and to the voxelwise fluences, which are normalized with respect to the number primary particles of the simulation. This effectively scales all fluence values relative to the maximum, ensuring that all fluences fall within the \([0..1]\) interval.
}

\subsection{Architectures}
In order to access the effectivity of our neural network architectures, we implemented each architecture's neural network model in three variants to handle the different provided parameters for each dataset. With this approach, for each architecture there is one model that can only handle X-ray tube rotations, one that can additionally handle varying spectra and one that can additionally handle varying tube distances.

In general, our models all estimate the relative topology of the spatial radiation field distribution conserving the relation of intensities between all voxels inside one specific radiation field. This is important, as the actual fluence scales between different settings of the beam would introduce a bias towards high X-ray tube currents and thus harm the training of the neural networks. On the downside, in a real-world application, this inherently requires our approach to run concurrently to at least one area dosemeter that reports dose rates at a known point in space. As these devices, especially those with real-time measurement capabilities, are known to exhibit a certain energy and thus spectrum dependence, one needs to correct them accordingly. This is specifically of importance in our targeted use case, as we aim to predict the scattered field around the patient, where the spectrum is not homogeneous and is difficult to predict. Thus, we decided to train all our networks not using the actual target measurand, \eg air kerma rate \(\dot{K}_{\mathrm{air}}\) free-in-air or \added{ambient dose equivalent} \(H^*(10)\), but on the components of those measurands as provided by RadField3D, which are the volumetric \textit{photon fluence} \fluence and the volumetric \textit{photon energy distribution} \(p(E_\mathrm{\gamma})\) composed of \(32\) bins and \(E_{max} = 150\,\mathrm{keV}\). In fact, the concurrent measurements of an area dosemeter are only needed to correct the fluence. Therefore, one can potentially use the estimated spectra to correct an area dosemeter measurement \added{for its energy and angle dependence} in turn.

\subsubsection{Fully-Connected-Voxelwise-Networks.}
The primary family of neural networks that we explored is the family of \aclp{FCNN}. Those networks are not only the most primitive, but also the most flexible ones. They have long been proven to be universal function approximators~\cite{hornik_multilayer_1989} and as such, given the right input representation, they can effectively learn to reconstruct structures of various kinds. More specifically, we were inspired by the \ac{NeRF} approach of Mildenhall~\etal\cite{mildenhall_nerf_2020} which is used for \ac{NVS}. We chose this approach, as the reconstruction of radiance fields especially by follow up works for the real-time rendering application, seems promising for the reconstruction of radiation fields in real-time. As we chose to not only predict a scalar dose rate per voxel, but the photon fluence \fluence together with the normalized local photon spectrum \(p(E_\mathrm{\gamma})\), this matches the \ac{NeRF} architecture, where the reconstruction of visual density and color of a given point is separated as well.

For the training process of our \acp{FCNN}, we used a cosine annealing~\cite{loshchilov_sgdr_2017} learning rate scheduler with \(\mathrm{eta}_\mathrm{min} = 10^{-6}\) together with an Adam optimizer using weight decay~\cite{loshchilov_decoupled_2019} of \(10^{-4}\) and \(\beta_1 = 0.9 \land \beta_2 = 0.99\). Additionally, we preceded our main learning rate scheduler by a linear warmup phase that reaches our target learning rate after \(1000\) steps.

\paragraph{\textbf{\ac{SRBFNet}}:}
\begin{wrapfigure}{r}{0.4\textwidth}
\centering
\begin{tikzpicture}[
    node distance=0.48cm,
    every node/.style={font=\scriptsize}
]

\node[input, font=\scriptsize] (pos) {Location\\$\mathbf{x}$};
\node[input, right=1.0cm of pos, font=\scriptsize] (glob) {Global params\\$\mathbf{g}$};

\node[draw, circle, minimum size=0.8cm, below=of pos, font=\scriptsize] (fourier) {};

\draw[thick]
    ($(fourier.west)+(0.15,0)$)
    plot[smooth] coordinates {
        ($(fourier.west)+(0.15,0)$)
        ($(fourier.center)+(-0.15,0.25)$)
        ($(fourier.center)+(0.15,-0.25)$)
        ($(fourier.east)+(-0.15,0)$)
    };

\node[below=-0.5cm of fourier, font=\scriptsize, xshift=1cm] {$\gamma(\mathbf{x})$};

\node[block, below=of glob, font=\scriptsize] (genc) {Global\\encoding \(\mathbf{g}_{\mathrm{enc}}\)};

\node[block, below=1.2cm of $(fourier)!0.5!(genc)$, font=\scriptsize] (mlp1) {MLP Block 1};

\node[fuse, below=of mlp1, font=\scriptsize] (fusion1) {Feature\\Fusion};

\node[block, below=of fusion1, font=\scriptsize] (mlp2) {MLP Block 2};

\node[fuse, below=of mlp2, font=\scriptsize] (fusion2) {Feature\\Fusion};

\node[draw, circle, minimum size=0.8cm, below=of fusion2, font=\scriptsize] (residual) {\textbf{+}};

\node[block, below=of residual, xshift=-1.4cm, font=\scriptsize] (out1) {Fluence\\Projector};
\node[block, below=of residual, xshift= 1.4cm, font=\scriptsize] (out2) {Spectrum\\Decoder};

\node[draw, circle, minimum size=0.8cm, below=of out1, font=\scriptsize] (fluence) {\(\Phi^{3}_{\mathrm{\gamma}}\)};
\node[draw, circle, minimum size=0.8cm, below=of out2, font=\scriptsize] (spectrum) {\(p(E_\mathrm{\gamma})\)};

\draw[arrow] (pos) -- (fourier);
\draw[arrow] (glob) -- (genc);

\draw[arrow] (fourier.east) -| (mlp1.north);
\draw[arrow] (genc) |- (fusion1);

\draw[arrow] (mlp1) -- (fusion1);
\coordinate (emergeMLP) at ($(mlp1.south)+(0,-0.2cm)$);
\coordinate (skipLeft) at ($(mlp1.west)+(-1.2cm,-0.48cm)$);

\draw[arrow] (fusion1) -- (mlp2);
\draw[arrow] (mlp2) -- (fusion2);
\draw[arrow] (genc) |- (fusion2);

\draw[arrow] (fusion2) -- (residual);
\draw[arrow]
    (mlp1.south) -- (emergeMLP) -- (skipLeft) |- (residual.west);

\draw[arrow] (residual.south) -- (out1.north);
\draw[arrow] (residual.south) -- (out2.north);

\draw[arrow] (out1) -- (fluence);
\draw[arrow] (out2) -- (spectrum);

\end{tikzpicture}
\caption{Schematic of the fully connected architecture used as the backbone for SRBFNet, SPERFNet and PBRFNet outputting a spectrum and a fluence for each voxel location.}\label{fig:SRBF}
\end{wrapfigure}

This neural network forms the basis for reconstructing the presented datasets. For each dataset, we present variations of this network. In general, \ac{SRBFNet} receives the same input parameters as the original \ac{NeRF} did: The voxel location in cartesian coordinates (\(x,y,z \in [0,1]\)), normalized per dimension, and a normalized direction vector, which in this case is the X-ray tube direction. The overall structure is shown in~\cref{fig:SRBF}, where global parameters \textbf{g} in this case is the encoded direction vector.

Our network begins by feeding the location vector through an encoder using frequency encoding \replaced{using}{composed of} sine and cosine functions with different periodicities which were initially introduced by A. Vaswani~\etal\cite{vaswani_attention_2017} for the transformer networks and reused for the original \ac{NeRF} implementation. Even though, the direction vector can also be represented by a simple frequency encoding, we instead used a spherical harmonics based encoding as the cyclic nature of a rotation angle can be better represented by that encoding. This was demonstrated for \ac{NeRF}-like networks by InstantNGP of T. Müller~\etal\cite{muller_instant_2022}. For efficiency reasons, we used the spherical harmonics implementation as provided by the tiny-cuda-nn package~\cite{muller_tiny-cuda-nn_2021} with version 2.0. For both encodings, we appended the raw input to the encoded vector. In contrast to the classical \ac{NeRF} application, our network does not aim to learn the scatter objects 3D geometry with visual densities for volumetric rendering. Thus, we skipped the pre-processing path where the model should optimize the voxel location representation towards the local density independently from the direction. We want the network to learn the implicit representation of the fluence distribution, which is similar to the visual density, but in our case is ultimately dependent on the radiation direction. Thus, we extracted the fluence as well as the spectrum from the last layer. In order to build a robust backbone for solving the radiation field estimation under more complex parameter spaces, we did not just concatenate the feature vectors that describe the global beam conditions, such as the view direction in \ac{NeRF}. Instead, we experimented with various feature fusion methods and finally compared the \ac{FiLM}~\cite{perez_film_2018} approach, which was originally used in \acp{CNN}, with a \ac{GMU}~\cite{arevalo_gated_2020} approach and as the baseline, with the simple concatenation of feature vectors as is done in classical \ac{NeRF} networks. Ahead of fusion both feature vectors, the encoded X-ray tube direction vector was passed through the global encoding block, a \ac{MLP} block that is similar to the \ac{MLP} processing the encoded location vector with one hidden layer and a model width that is equal to the width of the main path of the network.

The actual hyperparameters used for this and the extensions of this network, such as the frequencies of the frequency encoding, spherical harmonics degrees, model widths and feature fusion layers are reported in the results section as we determined them using hyperparameter tuning. Through all layers of the main path of this network, except for the last ones, we use \replaced{SiLU}{\ac{SiLU}} activation functions. \replaced[R1C7, R1C12]{Only in the case of the output activation did we use different activation functions. For the fluence, we used a different activation function depending on the range of the fluence normalization. For the case of a normalization in [0,1], we used a sigmoid activation function, while for a normalization in [-1,1], we used a gradient conserving clamping clamp, with: (sigma and clamping equation) For the gradient conserving clamping, the second term of the sum is executed detached from pytorchs autograd graph which results in a clamping that still has the original gradients attached. }{For the output layer, we used two different activation functions. For the fluence, we used the sigmoid activation, to keep the output between 0 and 1:}

\begin{equation*}
    \sigma(x) = \frac{1}{1+\mathrm{exp}(-x)}
\end{equation*}

For the local spectra, we use a simple histogram normalization \(\mathrm{norm}_{\mathrm{hist}}\) as the activation function, to guarantee normalized spectra for the losses.
\begin{equation*}
    \mathrm{norm}_{\mathrm{hist}}(x) = \frac{x}{\sum_{i=0}^{n-1}(x_i)}, n = 32
\end{equation*}

\ac{SRBFNet} was designed to be trained on the parameter space provided by \textbf{DS-01}.

\paragraph{\textbf{\ac{SPERFNet}}:}
This network is the first extension of \ac{SRBFNet}. It extends the global parameters \textbf{g} by the X-ray spectrum of the radiation source. Each radiation field file contains the \(150\) bin histogram of the X-ray tube output spectrum using a bin width of \(1\,\mathrm{keV}\). In order to improve the radiation field distribution and include the reconstruction of the local spectra, we encoded that spectrum into our network. First, we reduced the dimensionality drastically by bilinear resampling the bins of the histogram down to \(64\) bins. The reduced histogram was then passed through a \ac{MLP} with one hidden layer using a width of \(32\) neurons and the \ac{SiLU} activation function. After the first linear layer, we implemented a layer-normalization to reshape the activation distribution.

The two feature vectors of the X-ray tube direction and the X-ray tube output spectrum were then concatenated and combined by the global encoding block in~\cref{fig:SRBF} that previously processed the direction vector.

\ac{SPERFNet} was designed to be trained on the parameter space provided by \textbf{DS-02}.

\paragraph{\textbf{\ac{PBRFNet}}:} This neural network targets the limitation of the previous described variant, that is only able to learn a fixed, static beam distance. Therefore, this model adds an additional encoder to allow for the additional beam parameter introduced by \textbf{DS-03} while keeping the same overall structure from \ac{SPERFNet}. This requires one new scalar input parameter, that is the tube distance as the opening angle of the rectangular beam is directly dependent on the distance. We observed, that simply passing this scalar raw leads the neural network to ignore it. The previously used frequency encoding is not useful here either, as the input data is neither of high frequency nor periodical. Thus, we added a mini \acp{MLP} into the architecture, which has a width of \(16\) neurons and in an optimal way encodes the distance information for the global parameters vector \textbf{g}.

\ac{PBRFNet} was designed to be trained on the parameter space provided by \textbf{DS-03}.

\subsubsection{Convolutional-3DVolume-Networks.}
Instead of predicting every single voxel by itself, which is inefficient when the aim is to always predict all voxels, especially for fine resolutions, one can use \acp{CNN}. These networks rely on using convolution kernels with multiple channels that sweep over the image space and by that, encodes local information. Initially, these networks were only used for Computer Vision, but by transposing this operation, these networks can be used for generative purposes.

\paragraph{\textbf{Beam2Scatter U-Net}:}
Our Beam2Scatter (B2S) U-Net is in general a \ac{CNN}, that encodes the raw radiation field of the direct beam prior to the interaction with the patient geometry and from that estimates the scattered radiation field as it is induced by the patient. Therefore, its encoder-decoder architecture follows the 3D-U-Net description~\cite{cicek_3d_2016}, which itself is just an adaption to the third dimension of the original U-Net, as it was initially used for biomedical image segmentation~\cite{ronneberger_u-net_2015}. We adopted this structure and gave our network the fluence map of the direct beam as a one-channel input and let the network predict a \(33\) dimensional output with the first \(32\) channels for the spectra and the last channel for the fluence. Another addition is, that we introduced \ac{FiLM}~\cite{perez_film_2018} layers after each up-sampling step in the decoder path, to modulate the weights of the decoder depending on the encoded input spectrum of the X-ray tube, which is encoded the same way as in \ac{SPERFNet}. This way we can inject that information into the U-Net structure. The networks outputs had the prior introduced activation functions applied: \replaced[R1C7, R1C12]{clampGC}{sigmoid} and \(\mathrm{norm}_{\mathrm{hist}}\).
Thanks to the direct beam radiation distribution being the input of this network, it can be used directly with all three datasets. For \textbf{DS-01}, the spectra encoding and fusion by using \ac{FiLM} layers was disabled.

\subsection{Training pipeline}
To ensure that the models are all comparable, the training pipeline for all models was identical. Regardless, whether the model itself was capable of predicting partial or whole fields, the evaluation of the loss functions and the sub sequential update of the weights was performed in the same way. All models were trained for a maximum of \(200\) epochs, which could be prematurely aborted when the validation loss was not improving for the last \(10\) epochs. For the training loop, we used the pytorch lightning~\cite{Falcon_PyTorch_Lightning_2025} framework \added[R1C15]{in version 2.5.6} which implements the Cross-Validation policy with a final test using a dedicated validation and test dataset. The ratio of the split into training, validation and test dataset was \(70\,\%\), \(15\,\%\) and \(15\,\%\).
At first, a number of fields were loaded from the dataset that are transformed to pytorch tensors, preserving the channels and single layers, and as such they were batched together. We used a physical batch size of \(4\) full radiation fields. Multiple batches of those predicted and ground-truth fields were aggregated together to an effective batch size of \(64\) for training.
We applied data pre-processing to each field prior to the prediction. This pre-processing included not only the option of joining the direct beam and scattered field channels, but also normalizing the ground-truth and input tensors. Finally, the loss functions for each of the neural networks normalized outputs (fluence and spectra) are calculated. The metrics are calculated on the inverse-transformed data in the original data space.

The initial learning rates were estimated per neural network by using the pytorch lighting implementation of a learning rate finder, which follows the descriptions of L. N. Smith~\cite{LRFinder}. We limited it to search on the interval of \([10^{-2}, 10^{-4}]\) for \(250\) steps.
 \paragraph{Per voxel training:} For the \ac{FCNN} models, which operate per voxel, we added an optional extra step to the training pipeline, that is typical for \ac{NeRF} approaches. We refer to it as voxel location randomization, which is essentially a perturbation of the voxel position around its center with a maximum jitter of half the voxel's dimension. This was used as a data augmentation to stabilize training in \ac{NeRF} variants like InstantNGP~\cite{muller_instant_2022}.

\subsubsection{Loss functions}
Across all our models we used the same two loss functions, that we optimized concurrently: \(\pazocal{L}_{\fluence}\) the loss regarding the volumetric fluence \fluence and \(\pazocal{L}_{p(E_\mathrm{\gamma})}\) the loss regarding the local photon spectrum \(p(E_\mathrm{\gamma}) \in \mathbb{R}^{32}\) for \(\sum_{i=0}^{31} p(E_{\mathrm{\gamma}, i}) = 1\).
\begin{equation*}
\begin{alignedat}{2}
    \pazocal{L}_{\fluence}(t, p) \;&= \frac{1}{3}\Bigl(\mathrm{L_1}(t, p) + \mathrm{L_2}(t, p) + \bigl(1 - \mathrm{SSIM}(t, p)\bigr)\Bigr),\ & \mathrm{with}\ t,p \in \mathbb{R}
\end{alignedat}
\end{equation*}
\paragraph{}
\begin{equation*}
\begin{alignedat}{2}
    \mathrm{CumSum}_i(x) \;&= \begin{cases}
        x_i, & i = 0,\\[6pt]
        \mathrm{CumSum}_{i-1}(x) + x_i, & i > 0.
    \end{cases}\\
    \mathrm{Wasserstein}(t, p) \;&= \frac{1}{n}\sum_{i=0}^{n-1}|\mathrm{CumSum}_i(p) - \mathrm{CumSum}_i(t)|, & \mathrm{with}\ t,p \in \mathbb{R}^{n} \\
    \pazocal{L}_{p(E_\mathrm{\gamma})}(t, p) \;&= \frac{3}{10}L_1(t, p) + \frac{7}{10} \mathrm{Wasserstein}(t, p), & \mathrm{with}\ t,p \in \mathbb{R}^{32}
\end{alignedat}
\end{equation*}
With  \(t\) being the ground truth (target) and \(p\) being the prediction of the neural network. \added[R1C16]{A quick exploration of the parameter space had lead to the presented ratio between Wasserstein and \(L_1\)-Loss and was beneficial for achieving satisfactory results.} Here, \(L_1\) and \(L_2\) refer to the L1- and L2-Norms using mean aggregation to produce a scalar loss for higher dimensional inputs.

\subsubsection{Hyperparameter tuning}
For the hyperparameter tuning, we used the Optuna~\cite{akiba_optuna_2019} framework with version 4.5. For the tuning of each network, we defined the search space for possible parameters as listed in~\cref{tab:tbl_hyperparams_search}. The tuning was, in general, performed on every model we discuss: B2S U-Net, \ac{SRBFNet}, \ac{SPERFNet}, \ac{PBRFNet}. Anyway, not all of the parameters could be tested for each model. For instance, the \ac{SRBFNet} does not encode any spectra information, therefore the spectra encoding related parameters were not probed.
\begin{table}[!h]
    \caption{Probed hyperparameter spaces for each parameter of the models (B2S U-Net, \ac{SRBFNet}, \ac{SPERFNet}, \ac{PBRFNet}), where applicable.}
    \label{tab:tbl_hyperparams_search}
    \centering
    \begin{tabular}{lc}
    Model parameter & Parameter space \\
    \hline
    model width & [64, 96, 128, 192, 256, 384] \\
    location encoding frequencies & [10, 12, 14] \\
    direction encoding harmonics & [4, 6, 8] \\
    feature fusion & [Concatenation, \ac{FiLM}, ResFiLM, GMU] \\
    random voxel location & [true, false] \\
    encoded spectra dimensions & [16, 32, 64] \\
    \hline
    \end{tabular}
\end{table}

\subsection{Validation Metrics}
To assess the quality of the developed models, we need to analyze their performance on the test section of our datasets. As the distribution of voxel values inside our dataset are unique to our use case and we have special requirements as well, we did not aim for perceptual accuracy as in most other reconstruction tasks, we first needed to assess which metrics and which variants of metrics are applicable for neural network evaluation of this kind. In general, we measure our metrics over the air kerma rate free-in-air fields \(\dot{K}_{\mathrm{air}}\) of ground-truth and predicted fields using the following three metrics classes.

\paragraph{\textbf{\ac{SMAPE}}:} The first and most obvious kind of metric we employed is the \ac{SMAPE} which provides the mean relative error across all voxels of the test dataset. As the dataset is strongly biased towards low voxel values we introduced this metric in two variants: \(\mathrm{SMAPE}_{acc}^{x\,\%}\) for the metric applied over all voxels whose values exceed \(\max(K_\mathrm{air}) *(1 - \frac{x}{100})\) and \(\mathrm{SMAPE}_{acc}^{scatter}\), which applies the metric on all voxels between \(5\,\%\) and \(0.5\,\%\) of \(\max(K_\mathrm{air})\).

The \ac{SMAPE} is defined in the following, where \(\mathbf{P}\) and \(\mathbf{T}\) are two sets of voxels, present in two corresponding \(K_\mathrm{air}\) fields. \(\mathbf{T}\) is the set of ground truth voxels and \(\mathbf{P}\) is the set of predicted voxels:
\begin{equation*}
    \mathrm{SMAPE} = \frac{2}{n} \sum_{i=1}^{n}{\frac{|P_i-T_i|}{|T_i| + |P_i|}}
\end{equation*}
\begin{equation*}
    \mathrm{SMAPE_{acc}} = 1 - \frac{1}{2} \mathrm{SMAPE}
\end{equation*}
With \(\mathrm{SMAPE_{acc}}\) referring to the \ac{SMAPE} accuracy metric reported in this work.

\paragraph{\textbf{\ac{SSIM}}:} This metric is a well established approach to measuring the similarity in terms of structure between a target and a prediction image or volume~\cite{zhou_wang_image_2004}. This metric is commonly used for reconstruction tasks and expresses the structural similarity between target and prediction by an intuitive value between \(0\) and \(1\), where $1$ indicates perfect accordance between both structures.
\added{
\begin{equation*}
    \mathrm{SSIM}(p,t) = \frac{(2\mu_p\mu_t+C_1)(2\sigma_{pt}+C_2)}{(\mu_p^2+\mu_t^2+C1)(\sigma_{p}^2+\sigma_t^2+C_2)}
\end{equation*}
with \(\mu_x\) being the mean of the window \(x\), \(\sigma_x\) being the variance of the window \(x\), and \(\sigma_{xy}\) being the covariance matrix of the two windows \(x\) and \(y\). The \(\mathrm{SSIM}(p, t)\) is calculated for small windows \(p\) and \(t\) of the compared volumes \(\mathbf{P}\) and \(\mathbf{T}\), and the global SSIM is calculated by the weighted sum of all windows. \(C_1\) and \(C_2\) are calculated from the dynamic range of the compared volumes as stabilization constants to prevent terms from degrading to zero.
}

\paragraph{\textbf{\ac{GPR}}:} The \ac{GPR} within the scope of a gamma evaluation is a common metric in the field of radiotherapy radiation transport simulation validation~\cite{yu_analysis_2019}. It reflects the relative number of voxels that pass a certain criterion within a given radiation field. The metric variant we implemented follows the idea of the \ac{MADD} as described by Jiang~\etal\cite{jiang_dose_2006}. For our validation we used two different criteria. The strict criteria, which measures the number of voxels with a \ac{MADD} of \(3\,\%\) using a spatial resolution of \(4\,\mathrm{cm}\) and the more relaxed criterion, which measures the number of voxels with a \ac{MADD} of \(10\,\%\) and a spatial resolution of \(6\,\mathrm{cm}\). To calculate the \ac{GPR}, the field is shifted by the allowed \(\Delta_{d}\) in each dimension, respecting the extent of a single voxel. For \(\Delta_{d} = 4\,\mathrm{cm}\) this results in \(2\) shifts per dimension for the case of a voxel size of \(2\,\mathrm{cm}\). The minimum distance weighted error for each voxel across the applied shifts is used as the error for the scoring of the passing voxels, thus the error  \(\gamma(\vec{x})\) for a voxel in a voxel grid \(\mathbf{X}\) at position \(\vec{x} \in \mathbf{X}\) is defined as:
\begin{equation*}
    \gamma^{\Delta_d}_{\Delta_{\dot{D}}}(\vec{x}) = \min_{\vec{r} \in \mathbf{V}_{\Delta_d}(\vec{x})}\Biggl(\sqrt{\Bigl(\frac{\|\vec{x} - \vec{r}\|}{\Delta_{d}}\Bigr)^2 + \Bigl(\frac{\dot{D}_p(\vec{x}) - \dot{D}_t(\vec{r})}{\Delta_{\dot{D}}}\Bigr)^2} \Biggr), \mathbf{V}_{\Delta_d}(\vec{x}) = \{ \vec{v}\,|\, \| \vec{v} - \vec{x} \| <= {\Delta_d} \land \vec{v} \in \mathbf{X}\}
\end{equation*}
with \(\dot{D}_t\) being the target dose rate in \(K_\mathrm{air}\) at a given point and \(\dot{D}_p\) being the predicted one.

The reported metric \(\mathrm{GPR}_{p\,\%}^{d\,\mathrm{cm}}\) follows from that as:
\begin{equation*}
\mathrm{GPR}_{p\,\%}^{d\,\mathrm{cm}} = \frac{|\{\vec{x}\,|\, \gamma^{d}_{p}(\vec{x}) \leq 1 \land \vec{x} \in \mathbf{X} \}|}{|\mathbf{X}|}
\end{equation*}

\paragraph{\textbf{Spectrum Accuracy}:} As the spectra represent a special kind of information that can be used for purposes other than direct dosimetry, we evaluate them separately. Therefore, we use the \ac{IoU}, where we calculate the ratio between the combined area of ground truth and prediction spectra and the overlapped area between them. This accuracy is defined as in the following, where \(\mathbf{P}\) and \(\mathbf{T}\) are two sets of voxels in two corresponding spectra radiation fields. \(\mathbf{T}\) is the set of ground truth voxels and \(\mathbf{P}\) is the set of predicted voxels:

\begin{equation*}
    \mathrm{Spec}_{acc} = \frac{\sum_{i=1}^n{\min(P_i, T_i)}}{\sum_{i=1}^n{\bigl(P_i + T_i - \min(P_i, T_i)\bigr)}}
\end{equation*}
\section{Results}
In general, the qualities of the developed models by this research work are condensed in~\cref{tab:tbl_results}. Therein, each metric value belongs to a network, trained, tested and optimized in terms of hyperparameters on one of our datasets. The first dataset \textbf{DS-01}, was used to train our base \ac{FCNN} architecture called SRBFNet. We compared the results of that network against the implementation of the U-Net that we introduced previously as Beam-to-Scatter (B2S) U-Net. As this network does not need any architectural changes to be applied on multiple datasets, as its inputs are the raw primary beam layer of a radiation field together with the X-ray tube output spectra, we used this as a reference for each dataset and variant of our \ac{FCNN} architecture. We report the best metric values our networks achieved after we optimized the models hyperparameters by using Optuna as mentioned in the previous section. The specific found hyperparameters for each network are listed in~\cref{tab:tbl_hyperparams}.
\begin{table}[h!]
    \caption{Reported reconstruction metrics for each network in its optimal configuration of hyperparameters. All metrics were calculated from the test dataset section, grouped by used dataset. \added[R2C4]{For the training of the models a NVIDIA H100 GPU was used.}}
    \label{tab:tbl_results}
    \centering
    \begin{tabular}{cccccccc}
        \hline
        Dataset & Model & \(\mathrm{SMAPE}_{acc}^{90\,\%}\) & \(\mathrm{SMAPE}_{acc}^{scatter}\) & SSIM & \(\mathrm{GPR}_{3\,\%}^{6\,\mathrm{cm}}\) & \(\mathrm{GPR}_{10\,\%}^{4\,\mathrm{cm}}\) & \(\mathrm{Spec}_{acc}\) \\
        \hline
        \multirow{2}{*}{{DS-01}}
        & SRBF & \(\mathbf{96.4}\,\%\) & \(\mathbf{95.5}\,\%\) & \(\mathbf{0.961}\) & \(\mathbf{99.8\,\%}\) & \(\mathbf{99.98}\,\%\) & \(\mathbf{84.2}\,\%\) \\
        & B2S U-Net & \(87.1\,\%\) & \(64.0\,\%\) & \(0.868\) & \(84.9\,\%\) & \(99.0\%\) & \(72.5\,\%\) \\
        \hline
        \multirow{3}{*}{{DS-02}}
        & SRBF & \(90.0\,\%\) & \(83.3\,\%\) & \(0.923\) & \(96.8\,\%\) & \(99.8\,\%\) & \(59.5\,\%\) \\
        & SPERF & \(\mathbf{97.1}\,\%\) & \(\mathbf{96.2}\,\%\) & \(\mathbf{0.952}\) & \(\mathbf{99.7\,\%}\) & \(\mathbf{99.9}\,\%\) & \(\mathbf{86.2}\,\%\) \\
        & B2S U-Net & \(88.0\,\%\) & \(55.5\,\%\) & \(0.844\) & \(72.7\,\%\) & \(98.1\%\) & \(77.1\,\%\) \\
        \hline
        \multirow{2}{*}{{DS-03}}
        & PBRF & \(\mathbf{84.4}\,\%\) & \(\mathbf{84.8}\,\%\) & \(\mathbf{0.902}\) & \(\mathbf{87.2}\,\%\) & \(\mathbf{94.4}\,\%\) & \(\mathbf{86.7}\,\%\) \\
        & B2S U-Net & \(19.5\,\%\) & \(19.4\,\%\) & \(0.802\) & \(21.8\,\%\) & \(30.5\,\%\) & \(14.2\,\%\) \\
        \hline
    \end{tabular}
\end{table}

On \textbf{DS-01}, one can easily see, that the \ac{FCNN} architecture outperforms the U-Net architecture. Even though, the U-Net also learns the structure of the radiation field\added[R1C17]{, as reflected by the SSIM scores in~\cref{tab:tbl_results}}, the edges between primary beam and scattered radiation are not reproduced as sharply as the \ac{FCNN} does. \added[R1C17]{This can be concluded from the relation between SMAPE accuracy on both regions and the SSIM scores, which drops less than the accuracy.}
This observation can be reproduced for \textbf{DS-02}. Additionally, we intended to highlight the effectiveness of our spectra encoding for the expected air kerma rates \(\dot{K}_{\mathrm{air}}\). Thus, we trained the base \ac{SRBFNet} model, which cannot handle different output spectra of the X-ray tube, on \textbf{DS-02}. The effectivity of our spectra encoding can be seen by comparing \ac{SRBFNet} against \ac{SPERFNet}, where we get a nominal difference of \(12.9\,\%\) for \(\mathrm{SMAPE}_{acc}^{scatter}\) just from the missing spectra information. Furthermore, \added[R1C18]{when comparing SRBF on DS-01 with SPERF on DS-02 in~\cref{tab:tbl_results}}, we can conclude, that training our \ac{FCNN} architecture on a bigger dataset using more degrees of freedom, namely the X-ray tube spectrum, actually improves the \(\mathrm{SMAPE}_{acc}\), even though the spatial resolution \added[R1C18]{, according to the drop in the SSIM score,} got slightly, but not notably, worse.

For \textbf{DS-03} and \ac{PBRFNet}, we observed, that using models with a higher neuron count by increasing the width of our model per layer beyond \(192\) neurons, \eg \(256\) or even \(384\), the reconstruction capability increases by up to \(1.5\,\%\) regarding \(\mathrm{SMAPE}_{acc}^{scatter}\). A similar increase of the SSIM on the other hand side, and thus the overall shape of the radiation field, is invariant to a change of the model width beyond \(192\) across all our datasets and models. As the benefit of using broader models is not significant enough to sacrifice performance for it, we limited the hyperparameter selection to models with a maximum model width of \(192\).

\added[R2C5]{From~\cref{tab:tbl_timings}, we can see, that the use of \ac{GMU} feature fusion is doubling the inference durations and decreases the learning capabilities consistently over all datasets and \ac{FCNN} based models.
The use of \ac{FiLM} feature fusion, on the other hand, is faster and often consistently beneficial for the SMAPE accuracy metric over the scattered radiation field region. 
In general, an inference duration for a whole radiation field estimation of about \(20\,\mathrm{ms}\) was achieved by all our \ac{FCNN} based models, when used with concatenation or \ac{FiLM} based feature fusion. Please note, that the corresponding B2S-U-Net implementation achieved a mean inference duration of \(7.5\,\mathrm{ms} \pm 2.0\,\mathrm{ms}\) on the same datasets and hardware using \ac{FiLM} conditioning in the center layer.}

\begin{table}[h!]
    \caption{Used model configurations, where L denotes the number of frequencies used for the Fourier feature encoding of the voxel location, \(l_{max}\) denotes the maximum degree used for the encoding of the radiation direction by spherical harmonics and \(\mathrm{dim}(\mathbf{g}_{\mathrm{spec}})\) indicates the dimensionality of the encoded input spectrum. Model width refers to the number of neurons in one full layer of the \ac{MLP} main path.}
    \label{tab:tbl_hyperparams}
    \centering
    \begin{tabular}{ccccccccc}
    \hline
    Dataset & Model & Model Width & L & \(l_{max}\) & \(\mathrm{dim}(\mathbf{g}_{\mathrm{spec}})\) & Conditioning \\
    \hline
    \multirow{1}{*}{{DS-01}}
    & SRBF & 192 & 10 & 4 & \(-\) & FiLM \\
    & B2S U-Net & \(-\) & \(-\) & \(-\) & \(-\) & \(-\)  &  \\
    \hline
    \multirow{2}{*}{{DS-02}}
    & SRBF & 192 & 10 & 4 & \(-\) & FiLM \\
    & SPERF & 192 & 10 & 4 & 16 & FiLM \\
    & B2S U-Net & \(-\) & \(-\) & \(-\)  & \(-\) & FiLM \\
    \hline
    \multirow{1}{*}{{DS-03}}
    & PBRF & 192 & 14 & 4 & 16 & FiLM \\
    & B2S U-Net & \(-\) & \(-\) & \(-\)  & \(-\) & FiLM \\
    \hline
    \end{tabular}
\end{table}

\begin{table}[h!]
    \caption{Inference durations per field (\(50^3\) voxels) of each model, regarding a specific feature fusion method, using non-compiled pytorch models and a NVIDIA RTX 4090 GPU \added[R2C4]{in a Intel i9-13900K system with \(128\,\mathrm{GB}\) of RAM installed}. All models were using the following configuration with a model width of \(192\) and encoding parameters \(l_{max} = 4 \land \mathrm{L} = 10\) for \ac{SRBFNet} and \ac{SPERFNet}, with one exception for \ac{PBRFNet}, which was using L = \(14\).}~\label{tab:tbl_timings}
    \centering
    \begin{tabular}{cccc}
    \hline
    Model & Feature fusion & Inference duration per field & \(\mathrm{SMAPE}_{acc}^{scatter}\) \\
    \hline
    \multirow{4}{*}{{SRBF}}
    & Concat & \(20.7\,\mathrm{ms}\ \pm\ 2.8\,\mathrm{ms}\) & \(94.3\,\%\) \\
    & FiLM & \(20.6\,\mathrm{ms}\ \pm\ 1.9\,\mathrm{ms}\) & \(\mathbf{95.5}\,\%\) \\
    & ResFiLM & \(24.5\,\mathrm{ms}\ \pm\ 1.8\,\mathrm{ms}\) &  \(94.4\,\%\)\\
    & GMU & \(41.0\,\mathrm{ms}\ \pm\ 2.6\,\mathrm{ms}\) &   \(88.7\,\%\)\\
    \hline
    \multirow{4}{*}{{SPERF}}
    & Concat & \(21.8\,\mathrm{ms}\ \pm\ 2.6\,\mathrm{ms}\) & \(95.2\,\%\) \\
    & FiLM & \(21.8\,\mathrm{ms}\ \pm\ 2.6\,\mathrm{ms}\) & \(\mathbf{96.2}\,\%\) \\
    & ResFiLM & \(24.7\,\mathrm{ms}\ \pm\ 1.6\,\mathrm{ms}\) &  \(95.4\,\%\) \\
    & GMU & \(41.7\,\mathrm{ms}\ \pm\ 2.5\,\mathrm{ms}\) & \(93.7\,\%\) \\
    \hline
    \multirow{4}{*}{{PBRF}}
    & Concat & \(21.0\,\mathrm{ms}\ \pm\ 1.3\,\mathrm{ms}\) & \(79.2\,\%\) \\
    & FiLM & \(22.2\,\mathrm{ms}\ \pm\ 2.5\,\mathrm{ms}\) & \(\mathbf{84.8}\,\%\) \\
    & ResFiLM & \(24.3\,\mathrm{ms}\ \pm\ 1.9\,\mathrm{ms}\) & \(67.8\,\%\) \\
    & GMU & \(40.4\,\mathrm{ms}\ \pm\ 2.4\,\mathrm{ms}\) & \(40.9\,\%\) \\
    \hline
    \end{tabular}
\end{table}


\section{Discussion}\label{sec:dis}
The configuration of the best performing models during the hyperparameter tuning are listed in~\cref{tab:tbl_hyperparams}.
\replaced[R1C7, R1C12]{From those results, we can conclude, that for the fluence prediction, the same data range and activation function as used for \ac{NeRF}-like approaches is beneficial, which is using sigmoid \(\sigma\) together with a linear maximum normalization between \([0,1]\). Moreover, w}{W}hen additionally reviewing the table of~\cref{tab:tbl_results}, we are convinced, that using a model width of \(192\) and a location encoding with frequencies within a range of \(10\) to at max \(14\) is sufficient for the presented use case. Consistently over all datasets, the angular resolution of the beam needed to solve this problem was found to be low enough, that a harmonics count of \(4\) for the spherical harmonics encoding of the beam direction is sufficient. Further, two plain \ac{FiLM} layers for feature fusion, one at an early and one at a late position in the network, was proven to be most effective for this use case compared to using \ac{GMU} layers, Residual-\ac{FiLM} layers and feature concatenation. Even though, the impact of the \ac{FiLM} layers on the inference duration is measurable, it is small compared to the concatenation that is commonly used in the field of \acl{NVS}, as it can be seen in~\cref{tab:tbl_timings}. In that table, we concentrate on the differences regarding \(\mathrm{SMAPE}_{acc}^{scatter}\), as this metric is reflecting the accuracy on the most interesting part of the predicted volume for the computational dosimetry. That is because the staff, during an \ac{IR} procedure, will be mainly monitored in the \(\mathrm{SMAPE}_{acc}^{scatter}\) volume. From~\cref{tab:tbl_timings}, one can also draw the conclusion, that the use of \ac{FiLM} layers did not only achieve the best accuracies, but also the shortest inference durations compared to \ac{GMU} or Residual-\ac{FiLM}. Moreover, we demonstrated, that the radiation transport simulation can be learned by the same deep, but narrow structure that is used to learn light transfer as in the use case of \ac{NVS}. As the provided implementation does not use custom, native GPU codes like implementations in CUDA or OpenCL, its inference time does not enable real-time usage in the sense of the required \(8\,\mathrm{ms}\) to \(11\,\mathrm{ms}\) for VR- or AR-applications. Nevertheless, an inference time of about \(20\,\mathrm{ms}\) is suitable for interactive applications. After all, with the provided \ac{FCNN} models, state-of-the-art inferences times, that are suitable for the VR- and AR-application, should be achievable by implementing the network as CUDA or OpenCL kernels in the future, as has already been demonstrated by InstantNGP~\cite{muller_instant_2022}.
\section{Conclusion}
We have presented three datasets for training machine learning and especially deep learning algorithms on spatially resolved and physically accurate radiation field distributions as they could arise during \ac{IR} procedures. Additionally, we have implemented and compared different variants of fully connected and convolutional neural networks, following the architectural design of U-Nets and \ac{NeRF}-like architectures. 
\replaced[R1C21]{Thereby, we have proven \ac{NeRF}-like architectures to be efficient and capable of being used for radiation protection purposes and}{Our results indicate that \ac{NeRF}-like architectures are effective and suitable for radiation protection applications, and are therefore likely applicable to the} training of medical staff in the context of \ac{IR} procedures.
We have also pointed out, which design decisions are effective in this special context. Our presented datasets together with the presented findings on the architectural decisions for neural reconstruction networks create a profound foundation for enabling further research on this topic, \eg extending our or similar approaches by uncertainty handling in neural networks or incorporation of dynamic scatter geometry.

\section{Future work}
\added[R2C2]{We leave it as future work to implement the presented \ac{FCNN} using fused CUDA kernels to improve their performance. In doing so, it is expected that the networks will achieve inference times suitable for \ac{AR}/\ac{VR} rendering, as we have presented lightweight \ac{NeRF}-like architectures with two \ac{FiLM} layers. We demonstrated that these \ac{FiLM} layers only add minimal overhead. In general, this approach was already demonstrated for \ac{NeRF} variants like InstantNGP~\cite{muller_instant_2022}.}
\added[R2C6]{Further, we empathize the importance of public available datasets for training and comparing neural networks for dosimetry. Especially, more datasets with increasing realism are needed by adding beam collimation that is independent from the tubes distance from the isocenter or by adding various patient shapes and poses to the dataset generation process.}

\ifpreprint
\section*{Data availability}
All the datasets are available from \url{https://box.ptb.de/getlink/fi5etj6QwfD5PFgNqbaSo1VP/JML-2025}.
\fi

%
%
%
%




%
%

\ack{The authors gratefully acknowledge partial funding by the DFG under Germany’s Excellence Strategy within the Cluster of Excellence PhoenixD (EXC 2122, Project ID 390833453) and a DFG Research Grant (HU 2660/3-1, "Development of real-time capable methods for the simulation of photon radiation - using the example of quantitative dosimetry in interventional radiology", Project number 547148940).}

\ifpreprint
\newpage
\fi

\bibliographystyle{unsrt}

\bibliography{bibliography-formatted}

\end{document}